\documentclass[]{llncs}
\usepackage{graphicx}

\usepackage{tikz}
\usepackage{comment}
\usepackage{amsmath,amssymb} 
\usepackage{color}

\usepackage[accsupp]{axessibility}  

\usepackage{caption}
\usepackage{subcaption}
\usepackage{ulem}
\usepackage{lineno}

\begin{document}
	\nolinenumbers
	\pagestyle{headings}
	\mainmatter
	\def\ECCVSubNumber{16}  
	
	\title
	{A New Spatio-Temporal Loss Function for 3D Motion Reconstruction and Extended Temporal Metrics for Motion Evaluation} 
	
	
	\titlerunning{Laplacian Loss Function and Extended Temporal Metrics}
	%
	\author{Mansour Tchenegnon\inst{1}
		\and	Sylvie Gibet\inst{1}
		\and	Thibaut Le Naour\inst{2}
	}
	\authorrunning{Tchenegnon et al.}
	%
	\institute{IRISA, Univ Bretagne Sud, Vannes, France \and
	Motion-Up Company, Vannes, France 
	\email{\{mansour.tchenegnon,sylvie.gibet\}@univ-ubs.fr, tlenaour@motion-up.com}
	}
	\maketitle
	
	\begin{abstract}
		We propose a new loss function that we call Laplacian loss, based on spatio-temporal Laplacian representation of the motion as a graph.	This loss function is intended to be used in training models for motion reconstruction through 3D human pose estimation from videos.	It compares the differential coordinates of the joints obtained from the graph representation of the ground truth against the one of the estimation. We design a fully convolutional temporal network for motion reconstruction to achieve better temporal consistency of estimation. We use this generic model to study the impact of our proposed loss function on the benchmarks provided by Human3.6M. We also make use of various motion descriptors such as velocity, acceleration to make a thorough evaluation of the temporal consistency while comparing the results to some of the state-of-the-art solutions.
	\end{abstract}

	\section{Introduction}

	%
	%
	
	The 3D human pose estimation is a topic that has been extensively studied in recent years. The goal is to reconstruct the 3D skeletal pose from the image. One way to achieve this goal involves first to estimate the 2D joint locations from the image~\cite{Cao18,Rogez20}. Then, from these 2D joint locations, the 3D skeletal pose can be estimated. This process can be extended to video data. Given an input video or a sequence of 2D joint locations for each frame of the video, the objective becomes to compute the 3D joint positions for each frame. Working with video brings some advantages. In particular, adding temporal information can improve the learning of depth information~\cite{Mehta17b}. Therefore, new solutions using temporal features from the input data are the subject of recent research.
	
	%
	%
	Many deep learning methods learn from temporal information. Among these, previous methods based on Recurrent Neural Networks (RNN) have shown their efficiency but in return are very time consuming. Other more efficient solutions have then been proposed, based on convolutional neural networks~\cite{Chen21,Shi20}. In this case, the motion is considered as a temporal signal of joint positions on which convolution filters are applied. Some researchers propose to learn temporal features from adjacent frames through a convolution network, and then regress this information to estimate the central frame~\cite{Pavllo19}. This solution improves the pose estimation but the temporal consistency of the motion is not necessarily better since the poses are still estimated one by one. In theory, the more accurate the pose estimation per frame, the more continuous the reconstructed motion. However, sufficient accuracy to obtain a temporally plausible reconstructed motion has not yet been achieved. Given this fact, we propose to use a convolutional neural network, called CVM-Net, to reconstruct motion through sequence-to-sequence pose estimation. The goal then becomes a motion reconstruction task. In this paper, the idea is not to outperform the spatial accuracy of recent neural network approaches, but to propose an approach that better preserves the temporal consistency.
	
	Besides the design of this fully convolutional network, our main contribution is the definition of a new loss function for sequence-to-sequence pose estimation, called ~\textbf{the Laplacian Loss Function}. This function exploits a spatio-temporal representation of the poses sequence, inspired by the Laplacian graph $3D+t$ model~\cite{LeNaour13}.
	Indeed, for this kind of sequence-to-sequence neural network, using only the default \textit{Joint Position Loss} function (average Euclidean loss for each posture of the sequence), leads to the averaging of the posture errors over the whole motion sequence. This is achieved by filtering out extreme postures and by favouring those postures that are most representative of the training data. By focusing on the Laplacian representation of motion, we take into account both the spatial structure of each pose, constrained by the skeleton, and the temporal trajectories of the skeletal joints.	
	
	Our goal is to preserve the temporal consistency of the reconstructed motion first and then achieve an acceptable spatial accuracy. Therefore, we need adequate metrics for the evaluation of the neural network that consider the temporal characteristics of movements. On the spatial aspect, the Mean Per-Joint Positions Error (MPJPE) used in all state-of-the-art solutions is the best choice. Most approaches limit their evaluation to this metric since most of them focus only on achieving the best accuracy in reconstructing the joint positions. Very few approaches make a temporal consistency evaluation. For that, they use the Mean Per-Joint Velocity Error (MPJVE), that evaluates the estimated velocity error of the joints, computed on adjacent poses. This metric is a good start to evaluate the temporal consistency of a motion. However, motion characteristics are not limited to the velocity. We propose to extend the evaluation to the acceleration, so we propose a metric to evaluate the acceleration of the reconstructed motion, that is the \textbf{Mean Per-Joint Acceleration Error, MPJAccE}.
	
	In this paper, we present in Section~\ref{sec:arch} the generic network for sequence-to-sequence estimation that we propose and the reasons for our choice. In Section~\ref{sec:loss}, after presenting some traditional loss functions, we propose a new spatio-temporal loss function based on the Laplacian representation of the motion. Section~\ref{sec:metrics} presents the extended metrics that we propose to evaluate the temporal consistency of reconstructed motion. Finally in Section~\ref{sec:results} we present and discuss the results of our experiments, including an ablation study for the \textbf{Laplacian loss} and the evaluation results based on metrics computed from motion descriptors.
	
	\section{Related Work}
	In this paper, we address the issue of reconstructing motion from video through 3D human pose estimation.
	
	\subsection{3D Human Pose Estimation}
	3D human pose estimation consists in estimating 3D skeletal poses given an image or 2D joint locations. According to the type of data used as input, we have two main categories.
	The first category uses the image as input and directly estimates 3D poses. In this case, the methods compute 2D and 3D features, such as heat maps and other features (camera focal length, depth information) to estimate the final 3D poses. These methods generally involve two stages: a features detection stage followed by a 3D pose estimation~\cite{Yang18,Habibie19,Luvizon18,Wei21}. Yang et al.~\cite{Yang18} use a 3D estimation pose network and a pose discriminator to ensure that the estimated poses are plausible. Wei et al.~\cite{Wei21} use a framework to generate heat maps and bone maps in order to extract 2D pose hypotheses. They then use a pose regressor or a selection-based algorithm on these hypotheses to compute the final 3D pose.
	
	In the second category, one starts by estimating 2D joint locations in the image using a 2D pose estimator. From the estimated 2D joint locations, through various methods, it is possible to estimate the corresponding 3D poses. The main advantage of this approach is that it is more efficient on videos in the wild, due to the use of state-of-the-art 2D estimators. Some researchers propose lifting models~\cite{Martinez17,Biswas19,Chen19,Zhao19,Shimada20,Zou21}. Martinez et al.~\cite{Martinez17} propose an approach using consecutive linear layers to perform a 2D-to-3D joint positions regression. Combining a 2D-to-3D pose regression and a 3D-to-2D pose re-projection modules, Biswas et al.~\cite{Biswas19} use information in . Chen et al.~\cite{Chen19} present an unsupervised algorithm that lifts 2D joints to 3D skeletons. They show that adding random 2D projections and an adversarial network allows the training process to be self supervised using geometric consistency. Shimada et al.~\cite{Shimada20} decide to first estimate 3D pose from 2D joints locations,	and then make the estimated pose more realistic, through foot contact prediction and physics-based pose optimization. Zou et al.~\cite{Zou21} and Zhao et al.~\cite{Zhao19} represent the 2D joint locations as a graph structure and use a Graph Convolutional Networks to estimate the 3D pose from it. Azizi et al.~\cite{Azizi22} encode transformations between joints using the M\"{o}bius Transformation and propose a new light Spectral GCN to achieve state-of-the-art results. All these approaches focus on 3D pose regression and achieves great results in 3D pose estimation.
	
	\subsection{Motion Reconstruction}
	
	\subsubsection*{Network architecture}
	Many 3D pose estimators are currently proposed in the literature. Most of them only work on one image at a time. When receiving a video as input, they estimate the pose at each frame, and then directly concatenate the outputs. This way of reconstructing a motion does not take into account the temporal characteristics of motion. This leads to some unsteady movements in the results, and very few approaches have considered these effects~\cite{Mehta17,Pavllo19,Cai19,Shi20,Wang20,Xu20,Chen21,Dabral18}.
	Among them, Metha et al.~\cite{Mehta17} choose to infer the pose at time $t-1$ to estimate the pose at time $t$. Wang et al.~\cite{Wang20} represent the 2D skeleton input as a spatio-temporal graph and propose a Graph Convolution Network to predict 3D poses. Xu et al.~\cite{Xu20} choose to first estimate 3D poses and then use a trajectory completion framework to correct the sequence. More recently, Shi et al.~\cite{Shi20} propose a CNN approach coupled with a skeleton model to correct the spatial joint positions. In their solution, two independent CNN models are first in charge of estimating the sequence of rotations and the bone lengths to preserve some of the skeleton constraints.
	From these features, they apply the forward kinematics model to obtain the sequence of 3D poses.
	Another solution proposed by Chen et al.~\cite{Chen21} is to predict the length and direction of the bones throughout the sequence and compute the 3D poses from these. 
	Our approach is based on temporal convolution to estimate 3D pose sequence from 2D joint locations sequence (obtained from video using a state-of-the-art 2D pose estimator).
	
	\subsubsection*{Loss function}
	Since motion reconstruction refers to spatio-temporal data, it is necessary to have an appropriate loss function to ensure a better learning process. The loss functions used in 3D human pose estimation are insufficient for this task because they focus on spatial accuracy alone. 
	To solve this situation, some researchers propose new loss functions based on temporal characteristics of the motion. 
	Among them, some choose to calculate the loss function by using the first derivative, that is the velocity~\cite{Zhang22,Cai19}.   
	Wang et al.~\cite{Wang20} propose a loss function, called ~\textit{Motion Loss}, computed from a motion pose encoding space.
	They project the predicted and ground truth joint positions into this space and compute the difference between the two encoded information. This difference evaluates the quality of the reconstructed motion. Unlike them, we propose a spatio-temporal loss function as a solution to the learning process.
	
	%
	%
	
	\section{Deep Learning and motion temporal features}\label{sec:arch}
	The human pose estimation task consists in estimating with accuracy the joints locations of the skeleton. Therefore, it mainly focuses on the spatial aspect of the motion, whether it uses a single frame or multiple frames. Our aim is to preserve the temporal aspect of the reconstructed motion, while achieving acceptable joint position errors.
	
	\begin{figure}
		\centering
		\includegraphics[width=100mm]{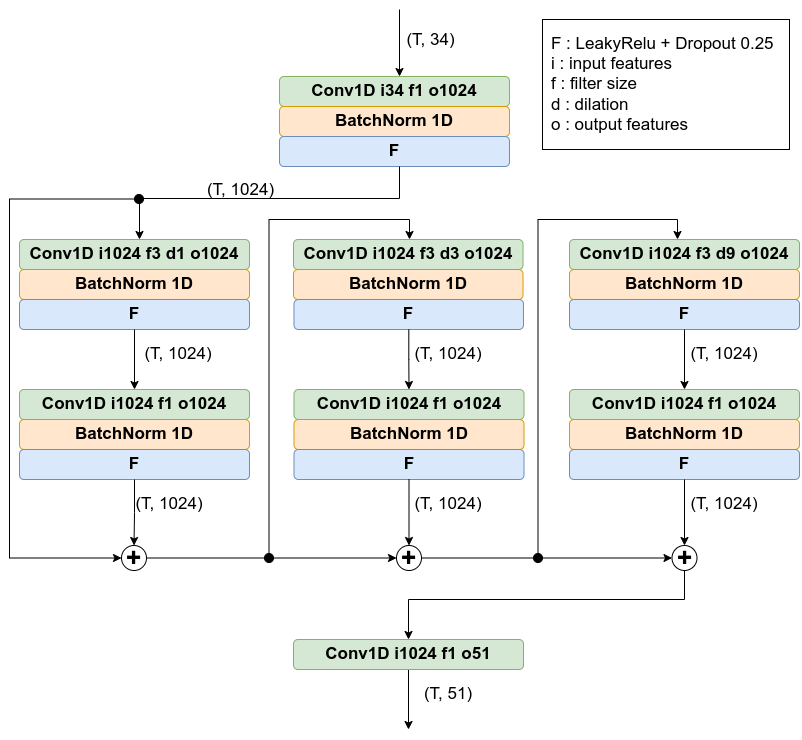}
		\caption{CVM-Net architecture. Temporal Convolution Neural Networks for motion reconstruction. This is a generic approach that combines 1D convolution to transform the 2D poses sequence into a 3D poses sequence. It also learn long duration relation between joints via dilation parameters}
		\label{fig:cmvnet-arch}
	\end{figure}
	
	Recent approaches prove that including temporal features of the motion improves the results in human pose estimation. A good technique to process temporal data is to use a RNN (Recurrent Neural Network). It allows to use information from previous frames to estimate the current one. However the computation is time and resources consuming. To overcome this drawback, an efficient and less consuming solution is to use Convolutional Neural Networks (CNN). In a survey conducted by Kiranyaz et al.~\cite{Kiranyaz21}, some studies on 1D convolution networks prove that they have a low computational complexity and are well-suited for low-cost and real-time applications. Besides, Arsalan et al.~\cite{Arsalan20} have also proven 1D convolution to be effective for trajectory-based air writing recognition. By considering a motion as a set of skeletal joint trajectories, 1D convolution is therefore a suitable choice to build our model.
	
	1D convolution applies filters using a time window. It takes as input $ b \times T_{in} \times C_{in} $ and output $ b \times T_{out} \times C_{out} $, where $ b $ is the batch size, $ T_{in} $ and $ T_{out} $ represent respectively the sequence lengths of the input and output, and $ C_{in} $ and $ C_{out} $ the channels or features for input and output. Moreover, we can use a dilated convolution to apply filters on non-consecutive frames to learn features at different time scales. Another advantage of the fully convolutional neural network is that it does not require a fixed sequence length and as a result can be easily generalized.
	
	Like many recent solutions, we have built our CVM-Net neural network to estimate poses through temporal convolution. By using multiple temporal convolution layers on frames in a time interval $ [t-\frac{\tau}{2}, t+\frac{\tau}{2}] $, of length $ \tau + 1 $, the accuracy of the the central frame estimation $ t $ is improved. But unlike most of these approaches, our solution estimates  multiple output frames ($ T_{out} $) from multiple input frames ($ T_{in} $), with $ T_{in}=T_{out} $.
	
	We used this neural network to study of the different functions we propose in this paper, namely the loss function and the motion evaluation metrics.
	
	%
	%
	\section{Loss Functions}\label{sec:loss}
	The loss function is one of the main part of a neural network. This is a key element in the training of neural networks, which indicates to the model its erroneous behaviour and brings possible corrections. The loss function is chosen according to the task at hand. Existing loss functions for human pose estimation and motion reconstruction focus on either the spatial aspect or the temporal aspect of the movement. Both are computed separately and them combined. This can be a limitation since each aspect, spatial or temporal, is considered independently. It is therefore necessary to find the appropriate coefficients of the linear combination while computing the global loss. In this section, we propose a loss function based on both the temporal and spatial characteristics of the motion.
	
	\subsection{Existing Loss Functions}
	\subsubsection{Joint Position Loss}
	This is the most commonly used loss function in a single frame pose estimation context. It computes an average distance between the joint positions of the ground truth poses and the estimated poses. Applied to multiple frames poses estimations it is defined as:
	
	\begin{equation}
	\mathcal{L}_{P}=\frac{1}{T} * \frac{1}{J} \sum_{t=1}^{T} \sum_{j=1}^{J} \Arrowvert  P_{t,j} - \overline{P}_{t,j} \Arrowvert _{2}
	\end{equation}
	
	where $P_{t,j}$ and $\overline{P}_{t,j}$ are respectively the 3D estimated position and the 3D ground truth position of joint $ j $ at time $ t $.
	This function, if used solely as loss function, works well for single frame pose estimation. But, when working on motion reconstruction, it is limited because it tends to average the joint positions loss over the whole sequence. The less represented poses in the motion can be biased by the more represented ones, affecting the overall motion reconstruction.
	
	\subsubsection{Motion Loss}
	Wang et al.~\cite{Wang20} propose a loss function as a distance in motion space. It is based on the encoding motion from a sequence of poses, by computing differential values between same joints at different time scales. It can be a subtraction, an inner-product or a cross-product. They encode both the estimated and the ground truth poses sequences. The loss is then computed between the encoded ground truth and reconstructed poses. They then combine this motion loss with the joint positions loss to compute an overall cost function defined by:
	
	\begin{equation}
	\mathcal{L} = \mathcal{L}_{P} + \lambda * \mathcal{L}_{M}
	\end{equation}
	
	where $ \mathcal{L}_{M} $ represents the motion loss, $ \mathcal{L}_{P} $ the joint position loss and $ \lambda $ a coefficient to apply on the motion loss.\\
	
	\subsection{Laplacian Loss}
	We propose a loss function for spatio-temporal features learning, which is based on the Laplacian representation of motion as a $3D+t$ graph, as defined by Le Naour et al.~\cite{LeNaour13}. The skeleton joints of the motion are considered as the nodes of the graph. The $3D+t$ graph is then obtained by i) first, connecting the joints to form the skeleton at each frame; these are the spatial edges. ii) We then create temporal edges by connecting joints between consecutive frames. Let's consider a motion of length $ T $ by a skeleton of $ J $ joints, and let $v_{j,t}$ be a node of the graph, representing the joint $j$ of the skeleton at time $t$. We create the temporal edges by connecting $v_{j,t}$ to $v_{j,t-1}$ and $ v_{j,t+1} $ which are the joints $ j $ of skeletons at time $ t-1 $ and $ t+1 $ respectively. The graph $3D+t$ is then defined by $G=(V,E_{S} \cup E_{T})$, with $V=\{v_{j,t}\}$ the set of all the joints, $ E_{S} $ the set of spatial edges and $ E_{T} $ the set of temporal edges.
	
	\begin{figure}[ht]
		\centering
		\includegraphics[width=100mm]{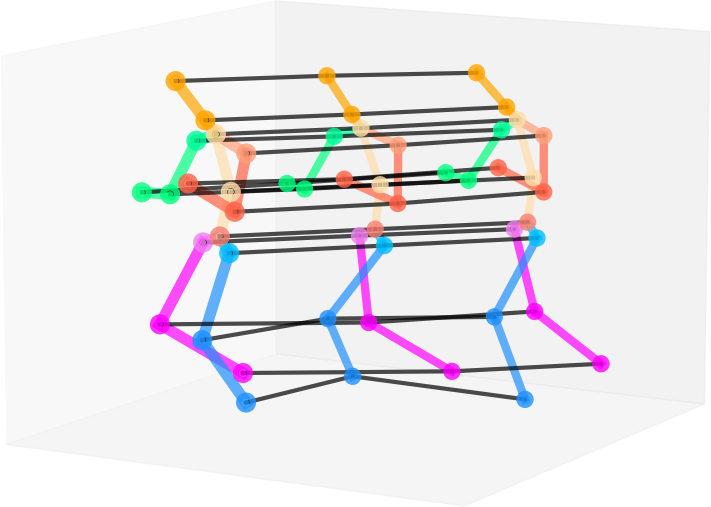}
		\caption{Graph $3D+t$ of a motion of length 3 for a skeleton of 17 joints. The black edges represent the temporal edges of $ E_{T} $ while the coloured edges represent the spatial edges of $ E_{S} $.}
		\label{fig:graph3dt}
	\end{figure}
	
	Using this spatio-temporal graph allows to define in an unified structure, the spatial and temporal relations between each joints of the sequence. From the graph 3D+t, we extract the matrix $ L $. It is a square matrix of dimension $ (N \times N) $ where $ N=T*J $ represents the total number of joints (the number of frames $ T $ in the sequence is multiplied by the number of joints $ J $ per skeleton). $w_{i,j}$ represents the weight attributed to each edge of the graph. In our case we apply uniform weights for both spatial and temporal edges.
	
	
	\begin{equation}\label{laplacian-weights}
	w_{i,j} = \left\{
	\begin{array}{ll}
	1 & \text{if } (i,j) \in E_S \text{ or } E_T \\
	0 & \text{otherwise}
	\end{array}
	\right.
	\end{equation}
	
	Let's define the matrix $\Delta$, so called \textit{differential coordinates matrix}, computed from the Laplacian matrix $L$ and the joint positions matrix $ P $ is of dimension $ (N \times 3) $:
	$\Delta = L * P$.\\
	$ \Delta $ is then of dimension $ (N \times 3) $.
	The matrix $ \Delta $ represents the differential coordinates of each joint relatively to its neighbours. The closer the differential coordinates of the estimated poses to those of the ground truth, the better the reconstructed motion. We can therefore compute an average distance error between the differential coordinates of the ground truth and those of the estimation. This distance represents our loss value computed as follows:

	\begin{equation}\label{eq:mpjpe}
	\mathcal{L}_{\Delta} = \frac{1}{N} \sum_{1}^{N} \Arrowvert \Delta^{gt} - \bar{\Delta} \Arrowvert
	\end{equation}
	where $ N=T.J $ is the total number of joints with $ T $ the sequence length and $ J $ the number of joints for a skeleton, $ \Delta $ is the matrix of differential coordinates computed from the ground truth and $ \bar{\Delta} $ the matrix of differential coordinates computed from the estimation.
	By training the neural network with such a loss function will allow to learn the connections between the joints locations it predicts, thus taking into account implicitly the skeletal structure and the temporal evolution of the joints.
	
	$ \mathcal{L}_{\Delta} $ computes the mean absolute error between the differential coordinates extracted from the ground truth joint positions and the estimated joint positions. This loss is based on a differential representation of the original output and does not consider the absolute joint positions. Therefore we combine it with the joint positions loss to compute an overall loss:
	
	\begin{equation}
	\mathcal{L} = \mathcal{L}_{P} + \alpha * \mathcal{L}_{\Delta} 
	\end{equation}
	
	where $ \mathcal{L}_{P} $ is the joint positions loss, $ \mathcal{L}_{\Delta} $ is the Laplacian loss and $ \alpha $ is the coefficient applied on the Laplacian loss.	
	
	%
	%
	\section{Evaluation Metrics for Temporal Consistency Performance}\label{sec:metrics}
	There are two existing metrics for evaluating human pose estimation. The main metric used in the state of the art is the Mean Per Joint Position Error (MPJPE, same formula as the Joint Position Loss~\ref{eq:mpjpe}). It evaluates the spatial accuracy of the models, using the average errors in estimating the joint positions. The lower the error, the better the accuracy. The Mean Per Joint Velocity Error, on the other hand, evaluates the temporal consistency by computing the average velocity error between the estimation and the ground truth. In the state-of-the-art, it is computed as defined in equation~\ref{eq:mpjve}. Temporal consistency refers to how close the reconstructed motion is to the ground truth in terms of smoothness, velocity, or acceleration.
	
	\begin{equation}\label{eq:velocity}
	v_{j,t} = P_{j,t+1} - P_{j,t}
	\end{equation}
	
	\begin{equation}\label{eq:mpjve}
	MPJVE = \frac{1}{T-1} * \frac{1}{J} \sum_{t=1}^{T-1} \sum_{j=1}^{J} \Arrowvert  v_{t,j} - \overline{v}_{t,j} \Arrowvert _{2}
	\end{equation}
	
	where $ v_{t,j} $ and $ \overline{v}_{t,j} $ represent the velocity vectors of joint $ j $ at time $ t $, respectively from the ground truth and the estimation.
	
	In order to extend the temporal consistency evaluation to other motion descriptors. Therefore we define $ MPJAccE $ metric (Mean Per Joint Acceleration Error in equation~\ref{eq:mpjacce}), which is based on acceleration.
	
	\begin{equation}\label{eq:acceleration}
	a_{j,t+1} = P_{j,t+2} - 2 * P_{j,t+1} - P_{j,t}
	\end{equation}
	
	\begin{equation}\label{eq:mpjacce}
	MPJAccE = \frac{1}{T-2} * \frac{1}{J} \sum_{t=1}^{T-2} \sum_{j=1}^{J} \Arrowvert  a_{t,j} - \overline{a}_{t,j} \Arrowvert _{2}
	\end{equation}
	where $ a_{t,j} $ and $ \overline{a}_{t,j} $ represent the acceleration vectors of joint $ j $ at time $ t $, respectively from the ground truth and the estimation.
	
	\section{Experiments}\label{sec:results}
	
	\subsection{Ablation Study}
	To evaluate the performance of our new loss function, we first set up training-test experiments using the neural network architecture proposed in section~\ref{sec:arch}. We use the same training environment for each session. In these experiments, we compare three configurations of loss functions.
	\begin{itemize}
		\item \textbf{CVM-Net} uses only the joint position loss $\mathcal{L}_{P}$ as cost function (baseline).
		\item \textbf{CVM-Net + $ \mathcal{L}_{M} $} uses a combination of the joint positions loss $\mathcal{L}_{P}$ and the \textit{Motion Loss} $ \mathcal{L}_{M} $ as proposed by Wang et al.~\cite{Wang20} in an overall function.
		\item \textbf{CVM-Net + $ \mathcal{L}_{\Delta} $} finally uses a combination of the joint positions loss and our \textit{Laplacian Loss} $\mathcal{L}_{\Delta}$ in an overall function.
	\end{itemize}
	
	\begin{table}[!ht]
		\caption{Comparison of the three loss functions: CVM-Net (baseline), CVM-Net + $\mathcal{L}_{M}$, CVM-Net + $\mathcal{L}_{\Delta}$, with the reconstructed errors MPJPE (under Protocol-1) and MPJVE. Protocol-1 computes the MPJPE from joint positions relative to the root joint (central hip) by aligning the root joints of both the estimation and the ground truth. a) Comparison of loss functions with MPJPE; b) with MPJVE. For both metrics, the lower the better.}
		\label{tab:abstudy}
		\begin{subtable}{\textwidth}
			\centering
			\caption{MPJPE comparison results in mm.}
				\begin{tabular}{ccccccccc}
					\hline\noalign{\smallskip}
					MPJPE & Dir. & Dis. & Eat. & Greet. & Phon. & Phot. & Pos. & Purch. \\
					\noalign{\smallskip}
					\hline
					\noalign{\smallskip} 
					CVM-Net & 85.25 & 129.41 & 109.17 & 101.22 & 117.17 & 137.31 & 86.12 & 293.37 \\
					CVM-Net w/ $\mathcal{L}_{M}$ & 83.67 & 107.73 & 118.72 & 95.51 & 113.39 & 131.98 & 82.64 & 221.00 \\
					CVM-Net w/ $\mathcal{L}_{\Delta}$ & \textbf{80.77} & \textbf{82.53} & \textbf{104.96} & \textbf{87.07} & \textbf{101.80} & \textbf{107.00} & \textbf{77.41} & \textbf{98.85} \\
					\hline
				\end{tabular}
				\begin{tabular}{ccccccccc}
					\hline\noalign{\smallskip}
					MPJPE & Sit. & SitD. & Smok. & Wait. & WalkD. & Walk. & WalkT. & Avg \\
					\noalign{\smallskip}
					\hline
					\noalign{\smallskip} 
					CVM-Net & 152.75 & 248.98 & 119.87 & 105.45 & 261.62 & 87.20 & 87.81 & 142.47 \\
					CVM-Net w/ $\mathcal{L}_{M}$ & 148.74 & 232.07 & 113.56 & 95.76 & 195.00 & 85.04 & 83.95 & 127.99 \\
					CVM-Net w/ $\mathcal{L}_{\Delta}$ & \textbf{137.33} & \textbf{178.99} & \textbf{103.43} & \textbf{84.37} & \textbf{104.32} & \textbf{79.16} & \textbf{76.30} & \textbf{100.62} \\
					\hline
				\end{tabular}
		\end{subtable}
		
		\begin{subtable}{\textwidth}
			\centering
			\caption{MPJVE comparison results in mm/frame}
				\begin{tabular}{cccccccccc}
					\hline\noalign{\smallskip}
					MPJVE & Dir. & Dis. & Eat. & Greet. & Phon. & Phot. & Pos. & Purch. & Sit. \\
					\noalign{\smallskip}
					\hline
					\noalign{\smallskip} 
					CVM-Net & 3.34 & 4.63 & 3.39 & 4.73 & 3.11 & 3.87 & 3.17 & 9.21 & 2.78 \\
					CVM-Net w/ $\mathcal{L}_{M}$ & 3.25 & 4.73 & 3.15 & 4.44 & 2.91 & 3.84 & 2.95 & 11.39 & 2.54 \\
					CVM-Net w/ $\mathcal{L}_{\Delta}$ & \textbf{2.88} & \textbf{2.94} & \textbf{2.59} & \textbf{3.58} & \textbf{2.29} & \textbf{2.79} & \textbf{2.60} & \textbf{3.55} & \textbf{1.94} \\
					\hline
				\end{tabular}
				\begin{tabular}{cccccccc}
					\hline\noalign{\smallskip}
					MPJVE & SitD. & Smok. & Wait. & WalkD. & Walk. & WalkT. & Avg \\
					\noalign{\smallskip}
					\hline
					\noalign{\smallskip} 
					CVM-Net & 4.92 & 3.15 & 3.55 & 6.86 & 5.71 & 4.84 & 4.50 \\
					CVM-Net w/ $\mathcal{L}_{M}$ & 4.64 & 2.87 & 3.29 & 6.83 & 5.01 & 4.22 & 4.42 \\
					CVM-Net w/ $\mathcal{L}_{\Delta}$ & \textbf{3.22} & \textbf{2.25} & \textbf{2.56} & \textbf{4.13} & \textbf{4.20} & \textbf{3.60} & \textbf{3.01} \\
					\hline
				\end{tabular}
		\end{subtable}
	\end{table}
	
	The study is achieved with the benchmark Human3.6M~\cite{Ionescu14} that contains millions of frames of captured data.
	For each configuration previously defined, we evaluate the MPJPE (through Protocol-1) and MPJVE using 2D joint locations ground truth as input (obtained from the benchmark). The evaluation consists of using the full dataset for the training-evaluation experiments. The dataset is split into a training set and a test set according to the subjects whose movements were captured (subjects 1, 5, 6 and 7 for training and subjects 9, 11 for evaluation).
	
	Table~\ref{tab:abstudy} presents the comparative results obtained with the different loss functions within the framework of Protocol 1 of the Human3.6M benchmark. Compared to the baseline, we observe an average decrease of $ 14.48 mm $ when we add the Motion Loss $ \mathcal{L}_{M} $ in the training process. In addition, the Laplacian loss $ \mathcal{L}_{\Delta} $ provides a significant improvement of $ 41.85 mm $ (average decrease in MPJPE). This shows that combining the joint positions loss with our Laplacian loss significantly improves
	accuracy. It also minimizes the velocity error by an average of $ 1.49 mm/frame $. This shows the efficiency of our spatio-temporal loss function that combines both the spatial relationships (between joints of the same skeleton) and the temporal relationships (between joints of consecutive frames).
	
\begin{figure}[ht]
		\centering
		\includegraphics[width=0.75\linewidth]{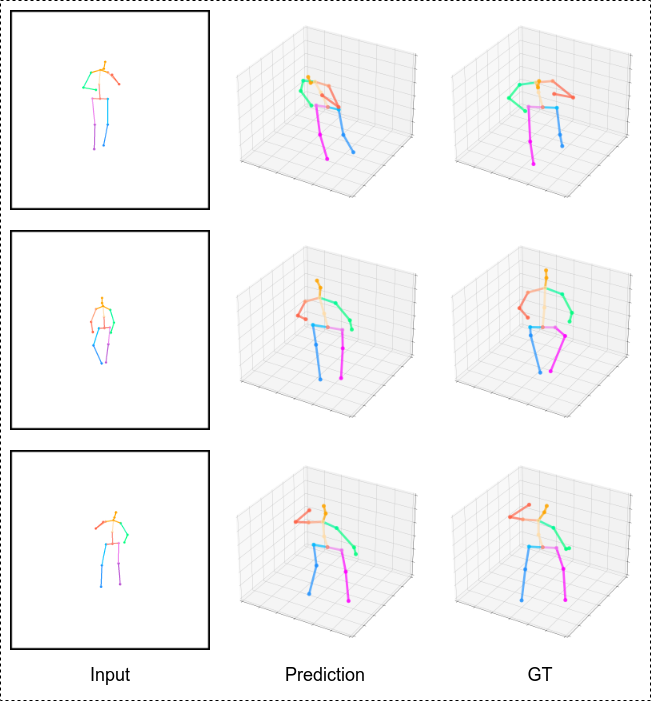}
		\caption{Visualisation results of CVM-Net + $ \mathcal{L}_{\Delta} $ on Human3.6m. On the left the input as 2D joint locations. On the middle our prediction and on the right the ground truth.}
		\label{fig:vis-results}
	\end{figure}

	Some visualisation results are displayed in Figure ~\ref{fig:vis-results}.
	
	\subsection{Evaluation Metrics Study}
	
	\begin{table}[!ht]
		\caption{Comparison with state-of-the-art solutions.
			a) Comparison with MPJPE under Protocol-1,
			b) MPJVE comparison,
			c) MPJAccE comparison. Best results are in bold.
			Legend : ($ \ddag $) sequence-to-sequence approach, ($ \dag $) sequence-to-pose approach with multi-frames as input and single frame as output, ($ \ast $) pose-to-pose approach, f=frame}
		\label{tab:full-evaluation}
		\begin{subtable}{\textwidth}
			\centering
			\caption{MPJPE comparison results in mm.}
			\begin{tabular}{ccccccccc}
				\hline\noalign{\smallskip}
				MPJPE & Dir. & Dis. & Eat. & Greet. & Phon. & Phot. & Pos. & Purch. \\
				\noalign{\smallskip}
				\hline
				\noalign{\smallskip}
				Shi et al.~\cite{Shi20} ($ \ddag $) & 45.48 & 51.28 & 49.43 & 51.91 & 52.58 & 66.46 & 50.59 & 48.46 \\
				Pavllo et al.~\cite{Pavllo19} ($ \dag $) & \textbf{33.88} & 43.99 & 44.28 & 48.96 & 44.62 & 65.80 & \textbf{32.79} & 55.12 \\
				Zhao et al.~\cite{Zhao19} ($ \ast $) & 38.62 & \textbf{43.08} & \textbf{35.89} & \textbf{40.15} & \textbf{40.85} & \textbf{50.14} & 42.56 & \textbf{40.40} \\
				
				\hline
				Ours(+$ \mathcal{L}_{\Delta} $) ($ \ddag $) &  80.78 &  82.50 &  104.44 &  87.03 &  101.16 &  106.88 &  77.43 &  98.24 \\
				\hline
			\end{tabular}
			\begin{tabular}{ccccccccc}
				\hline\noalign{\smallskip}
				MPJPE & Sit. & SitD. & Smok. & Wait. & WalkD. & Walk. & WalkT. & Avg \\
				\noalign{\smallskip}
				\hline
				\noalign{\smallskip}
				Shi et al.~\cite{Shi20} ($ \ddag $) & 55.90 & 64.25 & 53.79 & 52.84 & 58.85 & 49.99 & 48.25 & 53.47 \\
				Pavllo et al.~\cite{Pavllo19} ($ \dag $) & \textbf{45.61} & \textbf{48.09} & 57.29 & 47.09 & 45.16 & 43.30 & 46.67 & 46.84 \\
				Zhao et al.~\cite{Zhao19} ($ \ast $) & 47.81 & 56.47 & \textbf{42.20} & \textbf{42.25} & \textbf{42.29} & \textbf{33.39} & \textbf{36.00} & \textbf{42.14} \\
				
				\hline
				Ours(+$ \mathcal{L}_{\Delta} $) ($ \ddag $) & 136.73 &  178.48 &  102.98 &  84.33 &  103.65 &  79.24 &  76.39 &  100.34 \\
				\hline
			\end{tabular}
		\end{subtable}
		
		\begin{subtable}{\textwidth}
			\centering
			\caption{MPJVE comparison results in mm/f}
			\begin{tabular}{ccccccccc}
				\hline\noalign{\smallskip}
				MPJVE & Dir. & Dis. & Eat. & Greet. & Phon. & Phot. & Pos. & Purch. \\
				\noalign{\smallskip}
				\hline
				\noalign{\smallskip}
				Shi et al.~\cite{Shi20} ($ \ddag $) & 3.08 & 3.38 & 2.41 & 3.64 & 2.39 & 3.41 & 2.71 & 2.87 \\
				Pavllo et al.~\cite{Pavllo19} ($ \dag $) & 2.78 & \textbf{2.42} & 3.10 & 3.72 & 2.68 & 2.87 & 3.12 & \textbf{2.71} \\
				Zhao et al.~\cite{Zhao19} ($ \ast $) & \textbf{2.57} & 2.84 & \textbf{2.40} & \textbf{3.41} & \textbf{2.14} & \textbf{2.60} & \textbf{2.56} & 2.92 \\
				
				\hline
				Ours(+$ \mathcal{L}_{\Delta} $) ($ \ddag $) &  2.89 & 2.93 & 2.58 & 3.58 & 2.29 & 2.79 & 2.60 & 3.54 \\
				\hline
			\end{tabular}
			\begin{tabular}{ccccccccc}
				\hline\noalign{\smallskip}
				MPJVE & Sit. & SitD. & Smok. & Wait. & WalkD. & Walk. & WalkT. & Avg \\
				\noalign{\smallskip}
				\hline
				\noalign{\smallskip}
				Shi et al.~\cite{Shi20} ($ \ddag $) & \textbf{1.53} & 2.19 & 2.44 & 2.69 & 5.56 & 4.43 & 4.13 & 3.12 \\
				Pavllo et al.~\cite{Pavllo19} ($ \dag $) & 3.43 & 2.27 & 2.07 & \textbf{2.31} & \textbf{2.98} & \textbf{2.24} & \textbf{3.11} & 2.79 \\
				Zhao et al.~\cite{Zhao19} ($ \ast $) & 1.57 & \textbf{2.18} & \textbf{2.02} & 2.53 & 3.83 & 4.02 & 3.49 & \textbf{2.74} \\
				
				\hline
				Ours(+$ \mathcal{L}_{\Delta} $) ($ \ddag $) & 1.94 & 3.22 &  2.25 & 2.56 &  4.13 & 4.19 & 3.59 & 3.01 \\
				\hline
			\end{tabular}
		\end{subtable}
		
		\begin{subtable}{\textwidth}
			\centering
			\caption{MPJAccE comparison results in mm/f$ ^2 $}
			\begin{tabular}{ccccccccc}
				\hline\noalign{\smallskip}
				MPJAccE & Dir. & Dis. & Eat. & Greet. & Phon. & Phot. & Pos. & Purch. \\
				\noalign{\smallskip}
				\hline
				\noalign{\smallskip}
				Shi et al.~\cite{Shi20} ($ \ddag $) & 1.87 & 2.22 & 1.26 & 2.04 & 1.51 & 2.18 & 1.43 & \textbf{1.52} \\ 
				Pavllo et al.~\cite{Pavllo19} ($ \dag $) & 2.33 & 2.05 & 2.47 & 2.76 & 2.13 & 2.88 & 2.56 & 2.26 \\
				Zhao et al.~\cite{Zhao19} ($ \ast $) &  1.74 &  1.94 &  1.61 &  2.40 &  1.40 &  1.81 &  1.64 &  2.15 \\ 
				
				\hline
				Ours(+$ \mathcal{L}_{\Delta} $) ($ \ddag $) &  \textbf{1.12} &  \textbf{1.21} &  \textbf{1.03} &  \textbf{1.42} &  \textbf{0.98} &  \textbf{1.23} &  \textbf{1.06} &  1.83 \\
				\hline
			\end{tabular}
			\begin{tabular}{ccccccccc}
				\hline\noalign{\smallskip}
				MPJAccE & Sit. & SitD. & Smok. & Wait. & WalkD. & Walk. & WalkT. & Avg \\
				\noalign{\smallskip}
				\hline
				\noalign{\smallskip}
				Shi et al.~\cite{Shi20} ($ \ddag $) & \textbf{0.66} & \textbf{1.04} & 1.50 & 1.58 & 4.57 & 3.17 & 2.90 & 1.96 \\ 
				Pavllo et al.~\cite{Pavllo19} ($ \dag $) & 2.72 & 2.05 & 2.09 & 2.07 & 2.34 & \textbf{1.81} & 2.64 & 2.34 \\
				Zhao et al.~\cite{Zhao19} ($ \ast $) & 1.02 &  1.52 &  1.23 &  1.64 &  2.83 &  3.01 &  2.34 &  1.89 \\ 
				
				\hline
				Ours(+$ \mathcal{L}_{\Delta} $) ($ \ddag $) & 0.88 &  1.91 &  \textbf{0.88} &  \textbf{1.03} &  \textbf{1.91} &  1.93 &  \textbf{1.50} &  \textbf{1.33} \\
				\hline
			\end{tabular}
	\end{subtable}
\end{table}

Usually, human pose estimators are evaluated with the MPJPE metric  which calculates the average error of the joint positions for each pose. Very few are evaluated with the MPJVE metric, which calculates the error on the velocity of movement. In our case, since we focus on the temporal aspect of motion, we added  to MPJPE and MPJVE, a new acceleration-based metric, MPJAccE (see section~\ref{sec:metrics}).
We compare our approach with some state-of-the art solutions using these three metrics. We chose three different types of approaches based on their input and output settings (sequence-to-sequence, sequence-to-pose, pose-to-pose). First the approach of Pavllo et al.~\cite{Pavllo19} is a sequence-to-pose approach from which we derived ours. The second solution from Shi et al.~\cite{Shi20} is one of the best sequence-to-sequence approaches in motion reconstruction that makes use of forward kinematics. The knowledge of the skeleton constraints (angle limits, bones lengths) is embedded in this model. Finally, the solution from Zhao et al.~\cite{Zhao19} is a pose-to-pose approach that achieves good results on 3D human pose estimation.
Table~\ref{tab:full-evaluation} shows the results obtained during our evaluation. Our approach using $ \mathcal{L}_{\Delta} $, although far behind the state-of-the art results for MPJPE ($ 2.5 \times $ less accurate), achieves good results for MPJVE and MPJAccE. The MPJVE results show that our proposal is comparable to the other solutions, although it is a generic approach. We achieve the best results for the MPJAccE metric. 
Since our approach does not take into consideration the preservation of skeletal bone lengths, unlike other methods, it is normal that the MPJPE measurement is unfavourable to us. As for the MPJVE metric, which calculates the vector of changes in position (both in velocity and direction), it filters in a certain way the joint positions. The acceleration (second order derivative), which characterizes the variations of the velocity, reflects another measure of motion features that filters the velocity. The results on the two metrics MPJVE and MPJAccE show that our method better preserves the temporal characteristics of the original motion than other approaches.

\section{Conclusion}
In this paper we have presented a new spatio-temporal loss function based on the representation of the motion as a 3D+t graph. We have shown that this function improves both the spatial accuracy and the temporal consistency of 3D sequence-to-sequence pose estimation for motion reconstruction. We have used a temporal convolutional neural network for sequence-to-sequence pose estimation on a large scale dataset Human3.6M. Although this generic model does not challenge state-of-the art solutions on spatial accuracy with MPJPE evaluation, it has proven the efficiency of the Laplacian Loss in the spatio-temporal encoding of motion and the improvement of the temporal consistency. Moreover, such a solution is to be preferred in a situation where we are interested in the differential features (speed, acceleration), for the in-depth analysis of the movement for example.
Based on these preliminary results, our future work will integrate this Laplacian loss with other strategies -- including bone length constraints -- to improve the spatial accuracy while preserving the current temporal consistency. 
We will also integrate this loss function into the training of existing state-of-the-art methods in order to further validate its efficiency and improve the results, both on spatial and temporal aspects. Finally, we intend to propose a post-processing method based on the Laplacian representation to correct the results of the methods having obtained the best scores (spatial accuracy) and thus to obtain a better temporal consistency.

In this work we also have proposed a new protocol for evaluating temporal consistency in motion reconstruction through 3D sequence-to-sequence pose estimation. Velocity and acceleration measurements provide metrics that extend the classical position metric and allow to evaluate different solutions according to criteria related to the temporal quality of motion.

\clearpage
%
%
\bibliographystyle{splncs04}
\bibliography{camera-ready}
\end{document}